# Automated and Autonomous Experiment in Electron and Scanning Probe Microscopy

Sergei V. Kalinin,[1,a] Maxim A. Ziatdinov,[1,2] Jacob Hinkle,[2] Stephen Jesse,[1] Ayana Ghosh,[1,2] Kyle P. Kelley,[1] Andrew R. Lupini,[1] Bobby G. Sumpter,[1] and Rama K. Vasudevan[1]

[1] Center for Nanophase Materials Sciences and [2] Computational Sciences and Engineering Division, Oak Ridge National Laboratory, Oak Ridge, TN 37831

**Abstract**

Machine learning and artificial intelligence (ML/AI) are rapidly becoming an indispensable part of physics research, with domain applications ranging from theory and materials prediction to high-throughput data analysis. In parallel, the recent successes in applying ML/AI methods for autonomous systems from robotics through self-driving cars to organic and inorganic synthesis are generating enthusiasm for the potential of these techniques to enable automated and autonomous experiment (AE) in imaging. Here, we aim to analyze the major pathways towards AE in imaging methods with sequential image formation mechanisms, focusing on scanning probe microscopy (SPM) and (scanning) transmission electron microscopy ((S)TEM). We argue that automated experiments should necessarily be discussed in a broader context of the general domain knowledge that both informs the experiment and is increased as the result of the experiment. As such, this analysis should explore the human and ML/AI roles prior to and during the experiment, and consider the latencies, biases, and knowledge priors of the decision-making process. Similarly, such discussion should include the limitations of the existing imaging systems, including intrinsic latencies, non-idealities and drifts comprising both correctable and stochastic components. We further pose that the role of the AE in microscopy is not the exclusion of human operators (as is the case for autonomous driving), but rather automation of routine operations such as microscope tuning, etc., prior to the experiment, and conversion of low latency decision making processes on the time scale spanning from image acquisition to human-level high-order experiment planning. Overall, we argue that ML/AI can dramatically alter the (S)TEM and SPM fields; however, this process is likely to be highly nontrivial, be initiated by combined human-ML workflows, and will bring new challenges both from the microscope and ML/AI sides. At the same time, these methods will enable fundamentally new opportunities and paradigms for scientific discovery and nanostructure fabrication.

[a] sergei2@ornl.gov

## I. Introduction

Imaging provides the basis for multiple areas of science exploring nature from astronomical to atomic scales. Electron microscopy, including (scanning) transmission electron microscopy ((S)TEM), and associated spectroscopy techniques, such as electron energy loss spectroscopy (EELS), are now key tools for probing atomic scale structures and functionalities in inorganic solids and hybrid materials, polymers, and biosystems.[1-3] Scanning probe microscopy techniques ranging from Scanning Tunneling Microscopy[4] to the broad spectrum of functional force-based Scanning Probe Microscopies[5, 6] enable imaging and spectroscopies from atomic to mesoscopic scales in a broad range of environments.[5-8] Similarly, optical, X-Ray, and mass-spectrometry imaging methods are the mainstay of multiple areas in chemistry, astronomy, and medicine.

Beyond imaging and spectroscopy, both SPM and STEM can be used to manipulate matter on the nanometer scale. This includes examples such as electrochemical lithographies[9] and ferroelectric domain writing in SPM,[10, 11] and atomic manipulation and assembly in STM.[12-15] While relatively less recognized in the context of atomic-resolution electron beam techniques, the direct electron beam manipulation of structure[16, 17] and ferroelectric domains[18, 19] on the mesoscale and atomic level[20, 21] and even direct atomic manipulation[22, 23] and structure assembly[24] have also been recently demonstrated.

However, these highly visible results of STEM and SPM research should be contrasted to the experimental research paradigm that has remained essentially unchanged over the decades. Traditional microscopy research workflows include long times spent optimizing microscope performance, with examples such as tuning the (S)TEM or tip conditioning in STM. In many cases, this process relies on the specific expertise of the operator and more rarely includes automated stages or allows for quantified performance. Notably, while it is common to refer to a "good" or "bad" tip in STM, rarely does this assessment come with the associated quantitative measures. Similarly, while the STEM community has developed measures of microscope resolution that describe the localization of an electron beam, this quantification does not usually extend to the details of the beam profile. Note that for (S)TEM the probe can in principle be optimized with a calibration sample, and steps such as sample preparation and tilting are performed with the knowledge that microscope parameters are independently configurable. On the contrary for STM and related techniques, probe and sample conditioning are performed jointly, typically resulting in considerably longer optimization cycles and slower throughput.

With the microscope performance optimized, the imaging process typically includes multiple imaging and spectroscopy measurements, aiming to identify the regions amenable for scanning, identifying the potential regions of interest based on observed structural images and spectral data, and detailed investigation of these selected regions targeting "publication quality" data. The process is typically repeated based on available time or human endurance, thus concluding the experiment. In special cases, the microscope can be configured to stationary long-term operation, e.g., during current imaging tunneling spectroscopy (CITS) in STM.[25, 26]

Once the data is acquired, the analysis stage is typically performed away from the microscope with the implicit understanding that the specific region of interest will be unavailable in the future (unless extensive effort to make the fiducial marks has been undertaken), and microscope parameters will be reproducible only within certain limits. Furthermore, the data analysis and interpretation, which typically involve multiple interactions with theorists and domain experts and in certain cases results in new quantitative insights and discoveries, can be associated with significant timescales. The experimental result informs the human operator and affects future sample selection and microscope operation (i.e., contributes to what is understood as domain experience), and is disseminated to a broad scientific community in the form of archival publications, codes, and data, and personal communications.

Previously, we have explored the role that imaging data can play in the broad context of scientific research and discovery,[27] some of the specific issues associated with SPM,[28] and associated opportunities in atomic fabrication.[29-31] In half a decade since then, some of the enabling tools for data and code sharing and connections within scientific communities have emerged, including the code repositories on GitHub such as Pycroscopy[32], AtomAI[33], Nion Swift[34], Py4DSTEM[35], and data repositories such as Citrination[36], NoMAD[37], and Materials Innovation Network.[38] Similarly, the development of cloud-based services such as Google Colab now allow for seamless integration of the scientific publication, code, and data sources as implemented in Jupyter papers,[39] books and papers with code, etc. However, despite some early demonstrations of controlled and automated experiments in SPM[40] and (S)TEM[21] and a number of opinion pieces highlighting some of the general challenges and opportunities,[30] the general analysis of this field balancing human and ML/AI decision making and instrumental considerations have been lacking. These opportunities are explored here.

## II. General considerations

In the present context, an automated experiment is broadly understood as a computer-driven microscope operation, in which some or most decision-making functions are transferred from the human operator to the ML/AI control system. Part of this concept stems from the realization that many of the stages in the microscopy workflow are time consuming and monotonous, even though they might require a qualified operator. At the same time, in many cases the intrinsic latencies of the imaging system are much shorter than human decision making. For example, the image acquisition in many electron microscopy modalities can be well below a second, whereas human-based decision making is typically considerably slower than that. However, we note that the vast majority of microscope operations require a human operator that has both microscope-specific experience (how to operate it) and general broad context of physical, chemical, or biological knowledge (what to look at and for). Therefore, the analysis has to be made in the broader context of human knowledge, including how the object of study and specific region are chosen, what drives the decision making during the experiment, and what analysis is performed on the imaging data. We further disambiguate the automated experiment where human-based decision making is part of the operation and the role of ML/AI is to assist with the specific tasks at low decision-making latencies, from fully autonomous experiment where the human role is setting the prior parameters and the ML/AI runs the full experimental process without human intervention. This in turn brings forth the question of the priors in the ML process including past knowledge and previous experiments, inductive biases of chosen algorithms, and potential interaction with the external sources of information, e.g., models and data repositories.

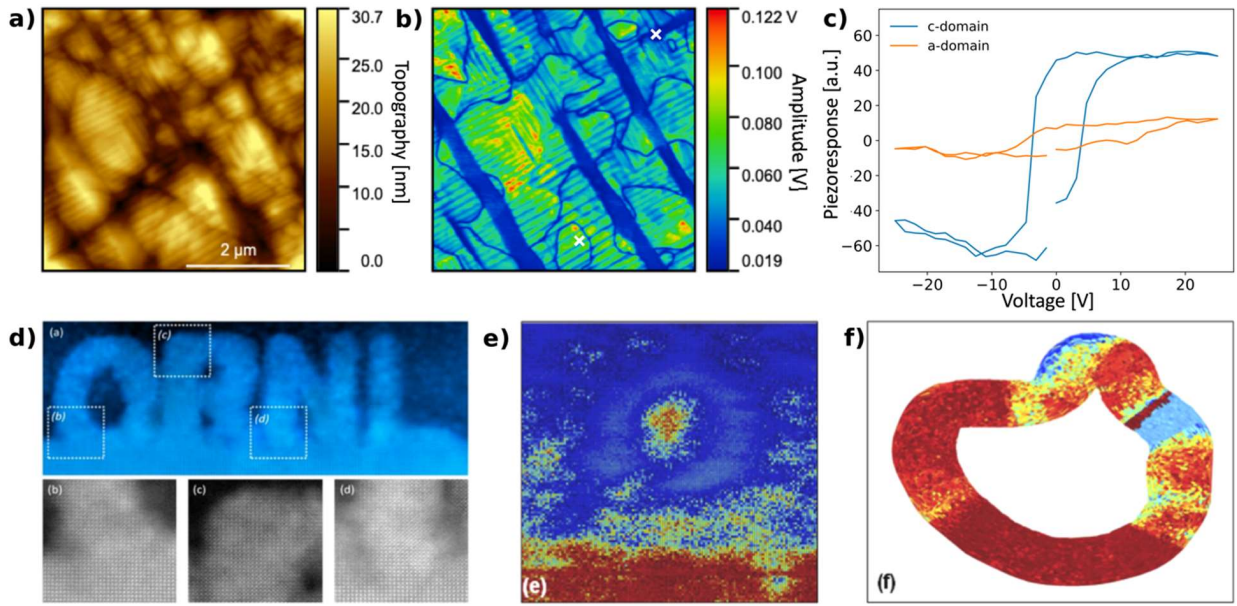

**Figure 1.** The motivation for automated experiment. Shown are (a) surface topography and (b) piezoresponse image of a lead titanate film, exhibiting a rich range of domain structures. Of interest for this material is the relationship between the ferroelectric-ferroelastic domain structures and out-of-plane polarization switching, as probed by (c) PFM spectroscopy. Performing these experiments on an equally spaced grid of points is highly inefficient, since most of the time will be spent on the regions with trivial domain structure. Similarly, polarization switching in PFM or (d) electron beam crystallization of matter on the atomic level necessitates control of the beam or the probe path (adapted with permissions from [20], (e,f) Example of the controlled scan path SPM realized by Ovchinnikov et al. (adapted with permissions from [40]).

Balancing this knowledge-based approach, we consider the practical aspects of instrumental operation, control, and performance. Unlike computational capabilities, the operation of mechanical systems is associated with intrinsic latencies and imperfections. Similarly, the signal at the detectors is associated with certain noise structures that may impose fundamental limits on how fast measurements can be taken or what signal to noise ratios can be achieved. Hence, the second set of considerations can be based on the analysis of the modifications and latencies of the instrumental scanning protocols and expansion of the traditional workflows. For example, autonomous experiment (AE) workflow can be based on a sequence of the classical full (or partial) rectangular scans and preconfigured spectroscopic measurements, i.e. emulating the human operator. Alternatively, the scanning can be performed on different trajectories, including pre-configured non-rectangular scans and ultimately even free-form trajectories. Similarly, the spectroscopic measurements can be based on preconfigured waveforms, utilize a range of waveforms while selecting between them, or use the freeform spectroscopy. In these, the key aspect is timing. For many SPM systems the image acquisition time is 1-10 min, whereas (S)TEMs can generate datasets on the ~1 s level and faster. Correspondingly, rapid feedbacks that control scan paths needs to operate well below these time scales, necessitating the development of edge computing capabilities.

Finally, an important issue is the stationarity of the systems. This includes the limitations of the experimental systems, including distortions and drift. Some of these can be corrected during the experiment (necessitating in turn additional stages in the experimental workflow), but some will be stochastic in nature. Similarly, in many cases microscopic studies will be performed on dynamic systems where materials structure and composition evolve as a function of temperature, gas atmosphere, or electrical stimulus. In these cases, AE design should consider the sample's changes. Perhaps even more important, in many cases the measurement itself can alter the sample structure. This can be a side effect of imaging, e.g., the beam damage in STEM or surface/tip degradation in SPM. For these cases, the experimental strategy generally attempts to minimize the damage. However, in many modalities change of the sample state is part of the measurement sequence, e.g., local hysteresis loop measurements in Piezoresponse Force Microscopy. Similarly, the sample changes can be harnessed to explore the broad chemical state of the system, performed control modification, and even accomplish atomic based fabrication. The goals and realizations of AE in this case has clearly different target then in exploratory microscopy.

Here, we organize the discussion following the operation workflows, and explore the ML options particularly paying attention to how human inputs and ML are balanced in terms of latencies and levels of decision making.

**III. Automating classical workflows: Sequential decision making**

Currently, most (S)TEM and SPM systems are based on rectangular scans for imaging and spectroscopic imaging, with the operator control being used to establish the regions of interest for scanning and adjust the field of view based on observations, as well as to select the regions for (typically longer-term and drift-prone) spectroscopic measurements. Typically, microscope scanning and correction protocols responsible for drift minimization and ensuring linearity are already optimized as a part of commercial development. Hence, we focus the discussion on the AE in this regime as closest to the existing technological paradigm.

The general paradigm in this case is the discovery and assessment for perceived interest and value of certain regions, with subsequent actions based on this assessment. The actions can include scanning at higher sampling to focus on a specific feature of interest, initiation of imaging or spectroscopy in a more complex and time-consuming mode, or modification. The object can be found based on morphological features or spectral responses. Selection can be based either on prior knowledge, novelty of the observed behaviors, or some combination of the two. Hence, in discussing the ML strategies, we put emphasis on discussing the prior knowledge available to a human operator and priors and inductive biases available for the ML algorithms.

**IV.1. Prior knowledge and inductive biases**

As mentioned above, any experiment is performed in the context of prior knowledge about the material and phenomena. Even prior to imaging *per se*, these considerations inform the sample selection and preparation. It is safe to assume that given the limited availability and access of most modern microscopes and complexity of sample preparation, experiments formulated on the spur of the moment will be exceptional. More generally, sample selection is based on a specific domain interest, hypothesis testing, or need for quantitative measurements. Whereas in many (and in the authors' experience, in most) cases the experimental studies reveal unexpected and serendipitous information, solid hypothesis-driven science is the basis of modern microscopy. Here, we assume that the choice of objects of study stays outside of the discussion and we refer the reader to more general works on artificial general intelligence[41] or AI/ML in science in general.[42]

However, this prior knowledge is also a significant factor in the microscopy experiment. Whereas the intrinsic microscopist's skill set includes operation and familiarity with the specific features and peculiarities of a given microscope, the whole process equally requires the knowledge of the material systems, identification of likely objects of interest and rapid decision making – which taken together generally constitutes the exploratory scientific research. Correspondingly, the key part of the AE is the clear separation of the prior knowledge available to ML system *vs.* that which is acquired during the experiment. Note that this knowledge can obviously include the weights of pretrained neural networks for image recognition and even the choice of network architecture. However, it also includes more subtle effects, for example choice of the acquisition functions in Bayesian optimization or reward functions in reinforcement learning (RL), balance of the exploration and exploitations strategies, inductive biases in both supervised and unsupervised learning methods, and choice of priors in Bayesian deep learning. Recognition of this prior knowledge is key for building successful machine learning based AE.

It is also important to mention that in principle both human and ML systems can seek additional knowledge during the experiment (active learning paradigm). For a human operator, it is common to query colleagues or stored information sources to address specific problems emerging during experiment as a part of troubleshooting procedure. It is considerably rarer to explore sample-specific information, e.g., to look up reference sources, during the experiment. In general, the operation of modern microscopes necessitates a certain focus on data acquisition, making the external information queries unusual. Comparatively, ML/AI in principle can seek new information during the microscope operation and adjust the experimental progression based on it, as these tasks can be done in parallel computationally. This information can be derived from correlative or generative analysis of experimental data, queries of the external data repositories, or running parallel modelling. Below discuss some of these opportunities.

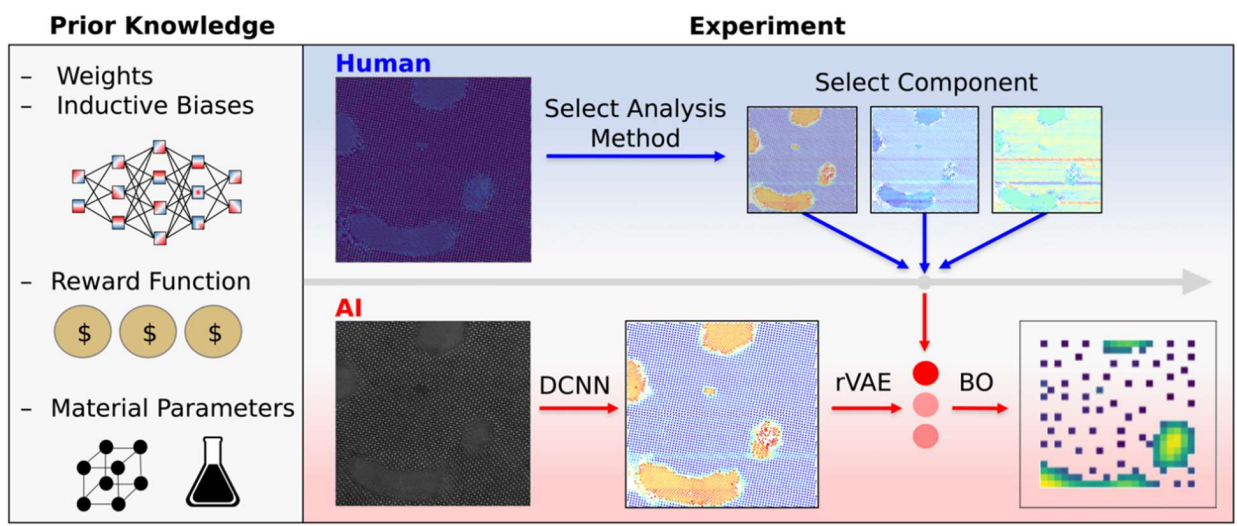

**Figure 2.** Example workflow for automated (but not autonomous) experiment. Here, the prior knowledge is incorporated in the form of pretrained network weights (for supervised learning), inductive biases (unsupervised learning), reward functions (reinforcement learning), and materials parameters. The human role is high-level slow decision making, e.g. choice of ML algorithm and selection of descriptors of interest. The AI role is fast low-level analysis that informs human decision making, and rapid analytics (reinforcement learning or Bayesian optimization) while operating the instrument. The STEM image is adapted from [43]

**IV.2. Machine learning in automated experiment**

Here we discuss the machine learning strategies in an automated experiment, namely the control and selection algorithm involved in the measurement workflow. One way to do so is based on the desired latency and the data on which the decision-making process is being made. We note that prior knowledge can be used explicitly or implicitly, e.g., in the form of the acquisition function selection in GP, combined functional form of data for multimodal and spectral images, choice of the reward design in RL, or implicit biases in self-supervised learning methods. Note that this analysis closely mimics a question of whether unsupervised machine learning methods such as variational autoencoders (VAE) can disentangle the data representations in the absence of inductive biases.[44-46] The special case here is the split of experiment into training and exploration stages, resembling classical NN networks or the Jeffreys prior[47] in Bayesian analysis.[48] In further discussion below, we aim to clearly disambiguate the role of prior knowledge and biases *vs*. AE operation.

### IV.2.1. Self-supervised experiments

The simplest case for AE are methods that control the experimental pathway based solely on the sample-specific data obtained during experiment and information on the microscope behavior. These include Gaussian Process (GP) and Bayesian Optimization methods, as well as reinforcement learning (RL) type decision algorithms. While not relying on direct human input, these methods nonetheless include prior knowledge in the form of the inductive biases, selection of the acquisition functions, reward structure in RL methods, or target performance. Below, we discuss these paradigms in detail.

### IV.2.1.a Gaussian Processes and Bayesian optimization

One of the key groups of algorithms that enable automated experimentation are the Gaussian Process (GP) regression[49-51] based Bayesian Optimization (BO). In general, GP refers to an approach for reconstructing a random function $f(\mathbf{x})$ over a certain parameter space $\mathbf{x}$ given the observations $y_i$ at specific parameter values $\mathbf{x}_i$. It is assumed that the observations, $y$, represent noisy measurements of the function, $y = f(\mathbf{x}) + \epsilon$, where $\epsilon$ is Gaussian noise. The values of the random function across its domain are related through a covariance function called a "kernel". The functional form of the kernel is chosen prior to the experiment, and kernel (hyper-) parameters are determined self-consistently from the observations ($y_i$, $\mathbf{x}_i$). The choice of the functional form of the kernel defines the physics of the explored phenomena. For example, the commonly used localized RBF and Matern kernels[52] can represent local correlations in the system, whereas spectral mixture kernels are well suited to systems with periodic structures.[53] In addition, the deep kernel learning method, which combines spectral mixture kernels and deep feedforward convolutional neural networks, can discover quasi-periodic patterns from sparse data and account for the non-stationary and hierarchical structure of the data.[54] When the domain is a regular image grid, GPR has a particularly convenient form, leading to efficient computation which has been used for image denoising and is applicable in microscopy contexts.[55-57]

The unique aspect of GP methods compared to other interpolation approaches is that GP analysis also provides quantified uncertainty, i.e. a function $\sigma(\mathbf{x})$ determined over the same parameter space $\mathbf{x}$ that defines the standard deviation of the expected values of $f(\mathbf{x})$. This naturally allows extending the GP approach towards automated experiments. Here, after the *n* initial measurements ($y_1, y_2, ..., y_n$) at locations ($\mathbf{x}_1, \mathbf{x}_2, ..., \mathbf{x}_n$) the function $f(\mathbf{x})$ and its uncertainty $\sigma(\mathbf{x})$ are reconstructed, and the location with maximal uncertainty, (i.e. a maximum of $\sigma(\mathbf{x})$) is chosen for a subsequent measurement. In this regime, AE minimizes the uncertainty over the predictions, corresponding to a purely exploratory strategy. In this exploratory strategy, the algorithm tends to focus on the edges of parameter spaces (e.g., scanning window boundaries) during the initial exploration steps, reflecting the large degree of "unknown" territory outside the explored interval. To avoid this tendency, the $\sigma(\mathbf{x})$ is can be multiplied by a mask that is constant within the parameter space and rapidly drops to zero at the borders.

Note that the GP variance maximization approach to AE mentioned above is essentially a form of adaptive optimal design of experiments (ADoE). Although the form of the uncertainty function in GPR, $\sigma(\mathbf{x})$, is independent of observed values *y*, it does depend on kernel hyperparameters. The kernel hyperparameters is chosen by maximum likelihood estimation, leading to an indirect dependence of chosen sample positions on observations *y*. Therefore, the adaptivity of this approach depends crucially upon repeated optimization of kernel hyperparameters, and upon flexibility in the class of kernel functions used. Note that the ADoE problem has been studied for GPR in other contexts, leading to methods that attempt to choose sample positions that are maximally informative to hyperparameter inference.[58]

In the BO methods, the exploration of the parameter space is guided by an acquisition function $a(\mu(\mathbf{x}), \sigma(\mathbf{x}))$ balancing the predicted functionality $\mu(\mathbf{x})$ and the uncertainty $\sigma(\mathbf{x})$. This practically means that regions of high uncertainty may not be explored if the optimization deems there to be little chance that probing such areas will result in a high target function value. Correspondingly, the definition of the acquisition function by a human operator captures the a priori knowledge about the system and allows for balancing between the pure exploration and exploitation strategies. The simplest acquisition function is a linear combination of the predicted functionality and uncertainty, where the choice of the coefficients in front of $\mu(\mathbf{x})$ and $\sigma(\mathbf{x})$ controls the balance between exploration and exploitation. Other choices for acquisition function include the probability of improvement, which indicates the likelihood of improvement over the current best measurement (e.g., maximum value of property of interest), and the size of the expected improvement (EI). Finally, one can define a custom acquisition function based on the specific research goals (if we know what particular behavior we want to target[59]) and/or allow for switching between different acquisition functions during the measurements.

While the BO methods have been well-known for decades, over the last 2-3 years they have received renewed attention in the context of automated experiments as a universal method to navigate low-dimensional parameter spaces. The broad range of BO optimization has emerged in the context of automated synthesis, where the parameter space is the composition space of the multicomponent phase diagram.[60-64] Similarly, BO has started to be extensively used in automated experiment in focused X-ray imaging and complex spectroscopies.[65-67]

The BO methods build on GPR foundations and can benefit from many advances therein; particularly the development of new kernel classes and sampling schemes. Flexible kernels such as DKL have been applied to BO[68] and kernel hyperparameter optimization is already part of some BO pipelines. However, future methods that integrate of ADoE objectives into acquisition functions in order to encourage efficient hyperparameter estimation during the BO process, may be successful in this context if they lead to increased adaptivity.

Until now, the BO methods used non-restrictive priors, assuming zero knowledge of the system. Correspondingly, when prior physical knowledge is available, the GP-based surrogate model for BO can be replaced with a probabilistic model of the system to make informed decisions about which combination of parameters to evaluate. A simple example in the case of STM would be accounting for a sample/tip electronic band structure (e.g., a presence of electronic band gap in a certain energy interval) when performing a BO search for optimal imaging parameters.

We further note that traditionally the BO methods are based on the definition of the acquisition function only, the maxima of which define the target parameter set for exploration. In most practical cases (relative short-range correlations in the image plane), the acquisition function tends to be shallow with multiple degenerate maxima. Correspondingly, development of the AE strategies should consider not only the acquisition function behavior, but also strategies for the sequence of exploration of the close maxima. Here, we refer to this as a pathfinder function, which balances expected microscope performance to yield the optimal sequence of exploration of close acquisition function maxima. For example, the pathfinder function can be based on proximity, impose a non-crisscrossing probe path, or favor probe motion in a specific direction.

Until now, we assumed that the input variables of the objective function in BO are independent, i.e. the causal structure existing among the input variables was not accounted for. The disregard for the causal relationships between the input variables may lead to suboptimal solutions, particularly for dynamically changing systems where a measurement is also an intervention (*e.g.* the electron beam in STEM is used to record an image but may also modify a material system). Recently, Aglietti *et al.*[69] introduced a new concept of causal BO that integrates observational and interventional data via a causal prior distribution and replaces a classical EI acquisition function with causal EI, which explores the possible interventions. As a result, in addition to the exploration-exploitation trade-off, the causal BO also balances the observation-intervention trade-off. However, we note that while the causal BO allows for a potentially significant decrease of the optimization cost/time, it requires a knowledge of a causal graph structure, which may not be always known in the actual experiments.

**IV.2.1.b. Predictability based learning**

The classical BO strategies rely on the uncertainties in the function interpolation as the basis of the acquisition function. However, often the observations yield structural or spectroscopic data sets with partially understood physical behaviors. This provides strong inductive biases that can be used for exploration. For example, analysis of structural data sets allows extracting structure-based descriptors such as polarization and unit cell volumes. The GP, ensembles of deep convolutional neural networks (DCNNs), or Bayesian Neural Networks can then be used to establish the correlative relationship between these physical descriptors, with the choice of feature and target descriptors being dictated by the inductive bias of the operator. With these in hand, the uncertainty in prediction can be used as a basis for exploratory studies.

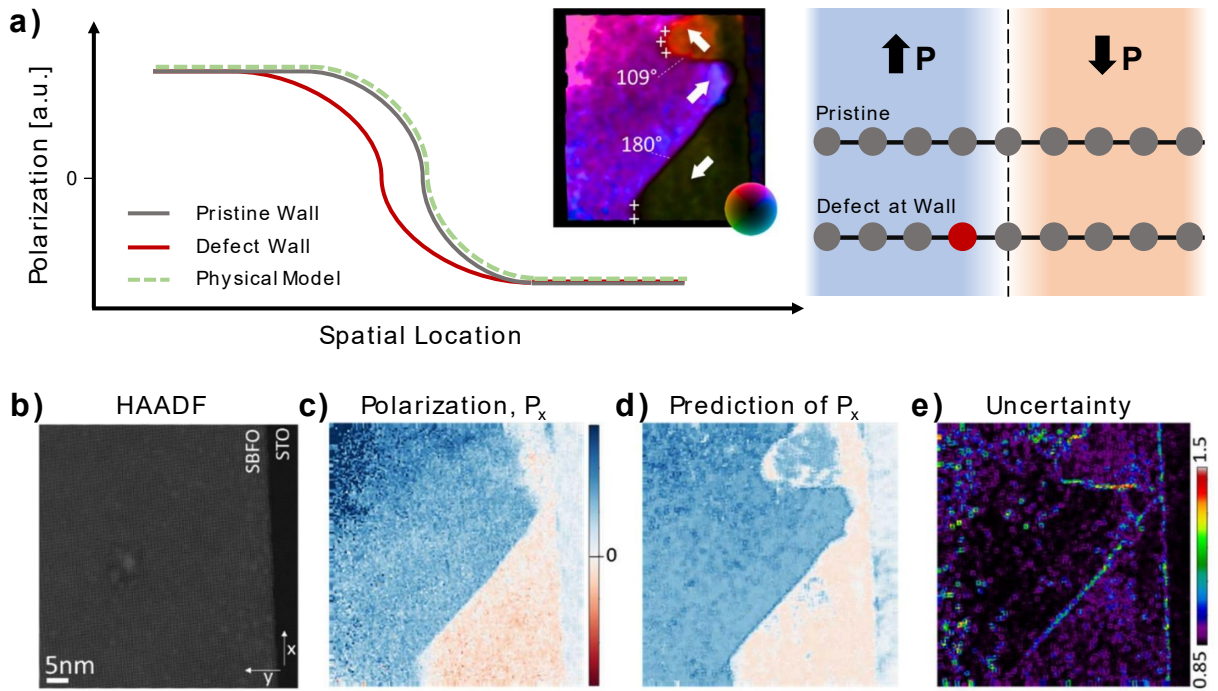

**Figure 3**. Predictability based exploration. Here, the uncertainty in predictions can be derived based on physical backgrounds, e.g., from the correlative relationship between locally probed physical descriptors. Schematically shown are (a) domain wall in the ferroelectric in with and without the defect center, which can affect the relationship between observed polarization value and molar volume. (b) STEM image, (c) ground truth polarization, (d) polarization predicted from structural descriptors, and (e) prediction uncertainty. The uncertainty map clearly delineates regions for more detailed studies associated with "unusual" behaviors. Adapted with permissions from [70].

Similar approaches can be used in cases where both structural and spectral observations are available. Typically, the structural information is available for images at a high sampling density, as exemplified by the STEM or STM images, whereas hyperspectral data is available over the same region but with lower sampling, e.g., EELS or 4D STEM in electron microscopy and CITS in tunneling spectroscopy. During the post-processing, these problems often give raise to the pan-sharpening and image fusion type problems, aiming to reconstruct the hyperspectral data at higher sampling.[71-73] Similarly, the encoder-decoder networks with the structural images as features and spectra as targets (*im2spec*) or *vice versa* (*spec2im*) can be used to establish the correlative relationship between the structural and spectral features.[74] The uncertainties in such predictions can also be used as a target for the automated experiment.

It is important to note that for optimal data representation, *im2spec* and *spec2im* networks should ideally be invariant with respect to the symmetries present during the imaging, *e.g.* allow for identification of rotated version of the same object, some translational invariances, scale invariances and so on. The approaches for introducing these via network construction or data augmentation have appeared only recently and will likely be an active area of research.[75-78]

Similarly, in the context of AE, it is important to compare cases where the deep learning networks are trained prior to the experiment, i.e., rely on past knowledge, or during the experiment. In case of the prior training, we note that in many cases a small deviation in the imaging parameters can result in out-of-distribution drift for the networks, degrading their performance. This problem can be addressed either via incorporation of the ensemble learning-iterative training (ELIT) methods[43] that allow to select the network suited for experiment from a distribution, transfer and active learning methods. This out-of-distribution drift can further be minimized by the transition from image-specific to materials specific descriptors. Alternatively, the networks can be trained using the data obtained at the initial stages of the experiment, and the thus trained network can be used to explore the uncertainties over the full AE sequence.

Similarly, for GP-based uncertainty evaluation, the disadvantage of the classical GP-BO approach is that that it does not "memorize" patterns/trends in the data and requires retraining from scratch after each new measurement, which may prevent its implementation in real time. Potential workarounds include the extension of the GP-BO approach to include the recently demonstrated meta-learning[79] and deep kernel transfer[80] approaches for GP.

### IV.2.1.c. Reinforcement learning

One of the key aspects to decision making in automated experiment is that for many cases, they require sequential decisions to be made without access to the result until many time steps after the initial decisions are made. In the standard Bayesian optimization routine, a new selection of point(s) results in new measurements, and this can then be used to retrain the surrogate model on which the optimization can be reformulated. This is in stark contrast to, for example, the goal of assembling atoms on surfaces to spell out the letters "IBM", which consists of many dozens, perhaps hundreds, of actions to pick up atoms and place them at specific locations.[12] The fact that the environment is stochastic (atoms can move, and the conditions of the experiment can change somewhat) can make automation in such environments difficult. These problems have recently been tackled by the advance of deep reinforcement learning (DRL).[81, 82]

In general, RL is a branch of ML that deals with *agents* performing *actions* in an *environment* with a goal to maximize a cumulative (discounted) reward. The agents act in the environment based on a given *policy*, usually parameterized by a neural network, that dictates which action to take given a particular state.[82] Note that "RL" refers to both the problem statement, as well as a class of algorithms designed to solve it.

More recently, the use of deep neural networks within traditional RL methods have attained high-profile success, with celebrated examples including the defeat of a Go champion by a DRL system developed by Deep Mind[83], as well as super-human performance in Atari games[84] as well as the "Defense of the Ancients" strategy game, the latter by OpenAI.[85] In many of these cases, the strategies learned by the agents are remarkably different from those of top players.[86] They have also shown steady progression in the area of robotics, for example with learning dexterous hand manipulations.[87]

Broadly two classes of RL exist, termed 'model-based' and 'model-free' RL. In a typical model-based RL setup, the agent interacts with the environment and learns a model of the dynamics (state transitions). Because repeated interactions with the environment are usually expensive, the agent performs a limited number of actions on the real environment, before updating the internal dynamics model, and then interacting again with the internal model to update the policy. This can be thought of as leveraging the data acquired in a more efficient manner, but one downside to model-based RL is that very often the policies learned from this process are optimal for the learned model, but not necessarily for the environment. Many of the more recent successes in RL utilize so-called model-free methods, which do not attempt to directly model the environment.

One of the key constructs in RL is the idea of a state value.[82] An RL agent following a policy may visit certain states and find that being in those particular states tends towards accumulating more rewards when following the policy through to the termination of the environment (or to a very long time). These states would be assigned a higher state value in the value function, $V(s)$, which is defined recursively as the immediate reward obtained at that state, plus the (discounted) reward at the next state, s' when following the current policy, i.e. $V(s) = r + \gamma V(s')$, where $\gamma$ is a discount factor that is between 0 and 1, and accounts for uncertainty in the actual state value. By visiting more states, the value function can be updated according to the actual returns (total accumulated rewards) of the agent. A similar construct is the use of the so-called action-value function or Q-function. When dealing with discrete action sets, it is often easier to work with the Q-function directly, which is defined $Q(s,a) = r + \max_a(\gamma\, Q(s',a))$, i.e. it shows the expected value of the state when taking a certain action (a). When this Q function is learned ('Q-learning') through the agent interacting with the environment, the Q values can be simply computed and the action that maximizes the Q value can be chosen at each state, thus defining the final policy. This approach was successfully applied to achieve super-human performance in Atari games,[84] for instance, and Q-learning has the advantage of being relatively data-efficient compared to other RL methods.

The alternative formulation for RL is that of direct policy parametrization. Q-learning is difficult to apply in practice for continuous action tasks, such as for example temperature control during a material synthesis problem, or electron beam dwell times. In principle these can be discretized for the sake of utility, but this may not be suitable in practice, essentially leading to very large action spaces. Stochastic policies can be represented directly by neural networks, and the policy gradient theorem can be utilized to update the parameters of the policy in the direction of maximizing the future discounted reward, as shown in Eq. (1). The policy defines a distribution of actions over states, and this distribution is usually parameterized, e.g. by a multivariate Gaussian distribution. Here, $J$ is the objective function, which is to maximize the cumulative discounted rewards $R$, which is a function of the trajectory $\tau$ through the system. These trajectories naturally come from following the policy $\pi_\theta$, i.e. they are sampled from the policy. Equation (1) indicates that the gradient of the objective function is obtained by collecting the returns from following trajectories and multiplying by the log probability of the actions that were taken. This (negative) 'loss' can then be back-propagated through the neural network in the standard manner to update the policy parameters, to make actions that increase the cumulative discounted rewards $R$ more likely. This defines the so-called 'Vanilla Policy Gradient' algorithm. The difficulty of this method is the high variance in R, which can be substantially reduced by subtracting an appropriate baseline from R that does not impact the gradient calculation. This baseline is often chosen to be the value function, $V(s)$, and incorporating this baseline essentially signifies the advantage that taking an action has over another, given the value of the present state, and is thus sometimes termed the 'advantage' function A. RL methods that use this function are often termed 'actor-critic' RL algorithms because they utilize an actor, that is the policy, and a 'critic', that determines the state value.[88] Both are updated during the course of the training. These methods are generally better for convergence than Q learning, but are data-intensive, and often only converge to local optima.[82]

$$\nabla_\theta J(\pi_\theta) = \mathop{\mathbb{E}}_{\tau \sim \pi_\theta} \left[ \sum_{t=0}^{T} \nabla_\theta \log \pi_\theta(a_t|s_t) R(\tau) \right] \qquad (1)$$

In most RL problems, the rewards given by the environment to the agent are remarkably sparse: in chess, the result of the game is only known at the end; in materials synthesis, roughness of the film can fluctuate considerably depending on the growth method, and final film properties will not be known until the synthesis is completed. The sparseness of the reward signal makes RL extremely data intensive, as the agent needs to encounter at least a few trajectories where some rewards are obtained. At the same time, insufficient exploration will lead to sub-optimal policies.

Multiple methods have been developed to attempt to better explore environments and lead to more robust policies. For example, the objective function can be modified to include an entropy term[89], so that it is not only rewards that are sought after, but rather, state exploration. This has been explored in multiple recent works (see Ref.[90] and refs therein) and can be assisted by parallelizing the exploration via use of multiple agents.[91]

As an alternative formulation, instead of consistently relying on rewards from the environment, the agent can act based on maximizing *curiosity* or *empowerment*. Curiosity can be defined (in one formulation) as the discrepancy between what an agent's internal model believes will occur to the state when an action is taken, vs. what actually occurs, in some featurized state space that is relevant for action selection.[92] It has been shown that utilizing curiosity driven RL can greatly accelerate learning on typical RL environments, because it leads to better exploration. An alternative signal is the empowerment signal[93], which can be formulated as the Kullback–Leibler (KL) divergence between the predicted successor state when a particular action is chosen, and the state when the actions are marginalized out. This effectively leads to actions which have the greatest effect on the environment to be rewarded. In both cases, these can (and generally are) added to the extrinsic reward signal coming from the environment itself. These methods promise to be extremely useful for AE because, in this case, it is not often possible to adequately construct reward signals if the goal is simple 'sample exploration'. Indeed, what we need instead is to find new states that are not predicted by the RL agent's learned model of environment dynamics. Moreover, it is expected that the learned policies can then be inspected to obtain an understanding about the physics of the underlying process, i.e., the specific features of the states which dictate the action, and the distribution of the actions themselves conditioned on different states.

When available, the most critical component of reinforcement learning is so-called 'reward shaping', i.e., the specific design of the reward function. Although the reward function can be sparse, discontinuous, and non-differentiable, poorly designed reward functions can lead to 'hacking' and trivial or useless policies. For example, imagine a task where the goal is to assemble a collection of dopants closer together using scanning probe manipulation for adatoms on a surface, or electron beam manipulation for dopants inside a material. If the reward is simply the distance between the dopants, then policies that destroy the lattice to make this possible might be expected and discovered, even though that is not desired.

### IV.2.2. Use of prior knowledge in AE

Alternatively, the AE workflow can utilize past knowledge, to some extent emulating the experience of a human operator. In this case, the operator defines both the target of the AE, and the prior knowledge and inductive biases available to the corresponding ML algorithm. This knowledge can be included in the form of trained classical- and DCNNs for feature selection for image-based feedback that define objects of interest and rely on the simplest form of prior experience in the form of correlative relationships encoded in the networks. Alternatively, this can be Bayesian priors on images and materials responses, etc. In general, this suggests that AE will be built based on complex workflows utilizing multiple levels of decision making.

**IV.2.2.a. General workflow**

A simple example of such workflows can be based on the combination of feature finding and Bayesian optimization as illustrated in Figure 4. Here, a pre-trained DCNN is used to identify the positions of the atomic units in the system, providing the coordinates and identities. A suitable encoder network can be used to transition from atomic coordinates to the latent variables describing local structures. Alternatively, graph analysis can be used to identify local molecular fragments. The atomic identities, latent variables, or *a priori* selected molecular fragments (e.g., isolated 5-7 rings in graphene) can be chosen as targets for specific intervention.

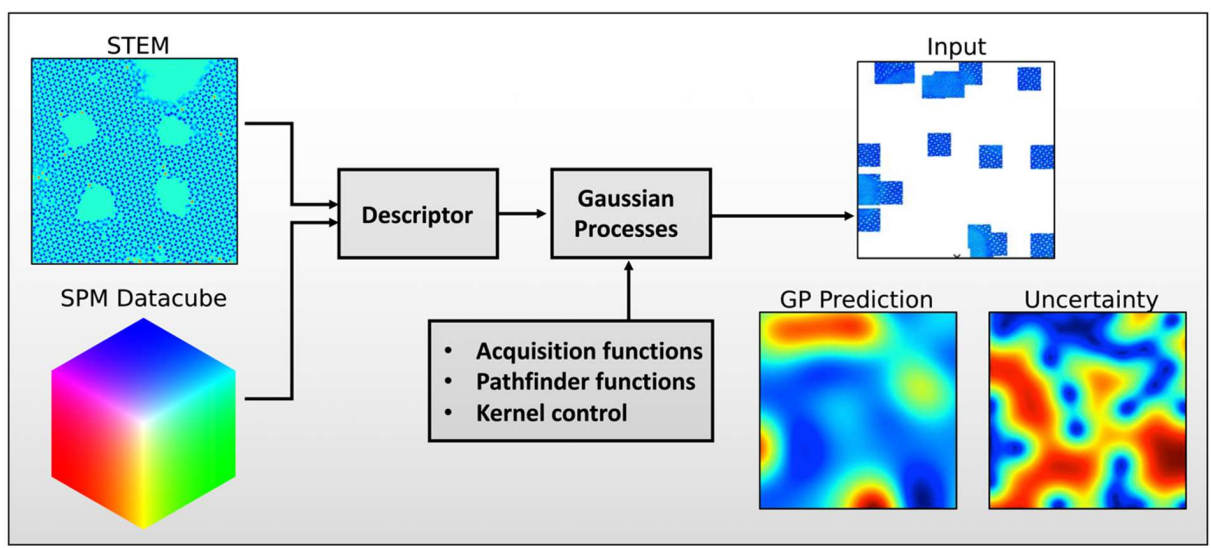

**Figure 4.** General BO workflow for structural and functional discovery in STEM or SPM. Structural or spectral data is reduced to a single or several scalar descriptors that are used to form the acquisition function in the BO process. Note that in many cases the classical acquisition function will be highly degenerate, and hence a pathfinder function that determines optimal measurement sequence on specific imaging system should be introduced.

Similar approaches can be used for more complex mesoscopic imaging data, e.g., polarization switching in piezoresponse force microscopy or many other methods. The network trains on the stack of images collected when applying a slow potential wave to get information on domain dynamics. These observations of domain dynamics allow establishing the latent variables (LV) that describe these images. Experimental strategies can then include trying to intervene on these latent variables (e.g. apply positive biases where LV is high, and negative where it is low), and explore effects of different interventions.

A simple example of such an approach is the FerroBOT concept in Piezoresponse Force Microscopy.[94] FerroBOT is based on line by line detection, where the changes in piezoresponse force microscopy phase when the tip crosses a ferroelectric domain wall are used to trigger the specific (predefined) voltage sequence that result in wall motion. This approach allows the systematic exploration of domain wall dynamics and also creating non-standard domain configurations. Further development of this concept includes development of simple image-based feedback using predefined features of interest.

Note that a key aspect of AE is the transition from the exploratory AE to active intervention, aiming to create target structures such as atomic assemblies or ferroelectric domain configurations via controlled modification of the material. This latter task is a considerably more complex problem, since the ML algorithm has to learn the (potentially stochastic) cause and effect relationships that control the beam and tip effects. This is a typical reinforcement learning problem as described above.

**IV.2.2.b. Dataset shift and out of distribution data**

Utilization of deep learning strategies have become common for categorizing natural images into different classes (cats, dogs, automobiles, etc.) or for performing a semantic segmentation of the images, that is, categorizing every pixel in an image into a particular class. Multiple public datasets, such as CIFAR and ImageNet, exist for benchmarking the newly developed deep learning networks. One of the characteristic aspects of these datasets is that each class usually contains thousands of examples with a high variability between them. By contrast, the problem of atom/particle/defect identification in SPM and STEM typically requires finding nearly identical objects while the image acquisition parameters can vary significantly between different experiments or between simulations and experiment. As a result, a neural network trained on a broad distribution of experimental parameters may not be able to recognize subtle (but physically important) differences between atomic features within a single dataset. Hence, the problem of the dataset shifts[95] and out-of-distribution samples becomes important.

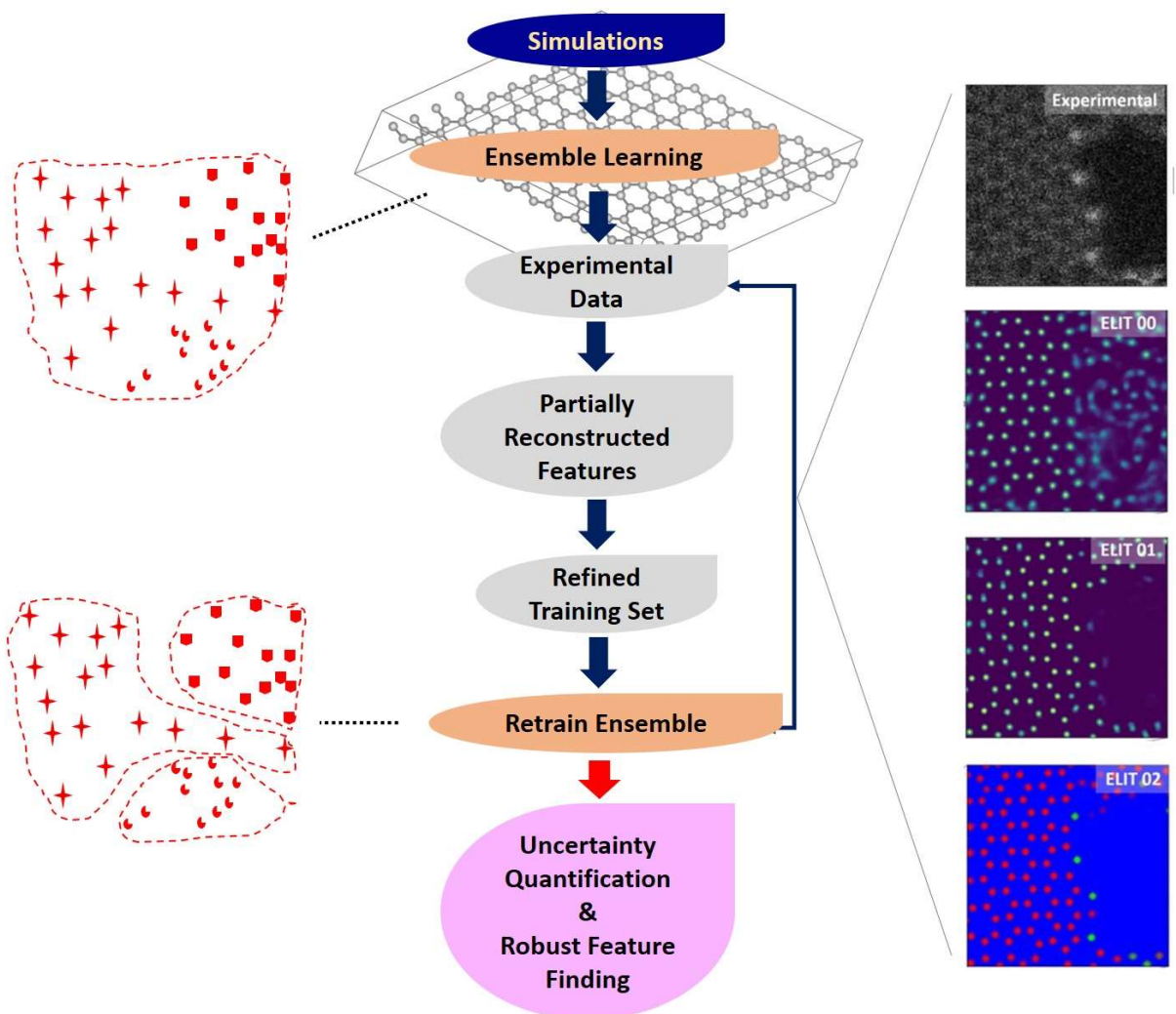

**Figure 5.** Illustration of the ELIT framework that starts with a neural network or an ensemble of neural networks trained on simulated data. This network(s) is usually "broad" enough to account for a variety of experimental conditions (noise, scan size and resolution, orientation, etc.) and atomic structures but may fail to detect a smaller variation in particular experimental settings or (mis-)identify physically impossible classes. Therefore, this network is used as a starting point (baseline) for iterative retraining procedure that allows focusing the network(s) attention on the classes present in the system, which could dramatically improve the classification accuracy.

One possible solution is utilization of ensemble learning and iterative training (ELIT) approach[43] that starts with an ensemble of models trained on simulated data (this allows avoiding manual labeling of experimental images) to identify the features present in a specific system with predictive uncertainties and then iteratively retrains the ensemble using the identified high-confidence features. An illustration of this approach is shown in Figure 5 using experimental STEM data on graphene (right panel) as an example. Here, the artifact-free high-quality prediction as well as associated uncertainties for each pixel (defined as dispersion in predictions of the ensemble models for each pixel; not shown) were obtained after just 3 ELIT iterations.

### IV.2.2.c. Inverse reinforcement learning and learning from human expert

One of the drawbacks of RL is the need to carefully tailor the reward function, as described in the previous section. For some problems, this is straightforward, but often reward functions will contain multiple components, necessitating substantial tuning, which is undesirable for a method that already incurs large and oftentimes prohibitive computational cost. Additionally, in many cases it may be difficult or perhaps even impossible to write an appropriate reward function. For example, consider the task of cleaning up a messy room. It is very easy to *demonstrate* how to do it, but much harder to numerically define such tasks with easy-to-understand parameters.

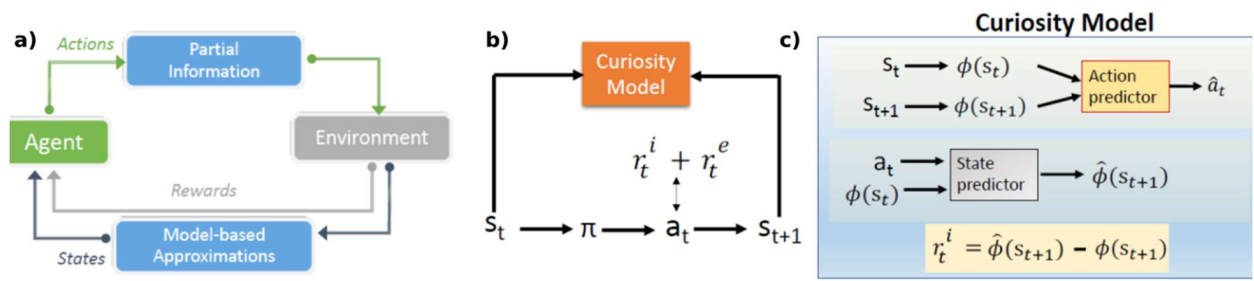

**Figure 6. (a)** General setup of reinforcement learning. In RL, an agent takes actions within an environment, usually based on partial information. The environment returns rewards, usually in some sparse fashion. The sparsity of the reward is at the heart of most of RL's sample inefficiency. One method to improve sample efficiency is to utilize intrinsic curiosity as an additional reward. **(b)** Curiosity-based RL. Here, the agent observes a state $s_t$ and samples an action from the policy π. This results in a transition to the next state, $s_{t+1}$. Meanwhile, the agent will receive any reward from the environment ($r^e$) and an intrinsic curiosity reward, $r^i$. The latter are given by the curiosity model, which is shown in **(c)**. In the curiosity model, two different models, parameterized by neural networks, are trained. One predicts the action based on two sequential states after featurization, and the other is an inverse that predicts the (featurized) next state. The difference between the featurized state prediction at time $t+1$ and the (featurized) observed state, $s_{t+1}$, is used as the intrinsic reward signal.

To tackle this problem, inverse reinforcement learning (IRL) is formulated to infer reward functions from observations of agents performing tasks in an environment.[96] In our case, this may be, for instance, a human expert demonstrating how to adjust multiple parameters on a microscope. The goal of IRL is to model the reward function from such demonstrations, or at least collect trajectory references, which can then be used to apply standard RL to leverage the expert demonstrations to learn a policy. As an example, IRL was recently used to model the reward function (using a deep neural network), trained on expert demonstrations playing Atari games.[97] These rewards were then used to train a deep Q network (DQN) agent that achieved superhuman performance without using game rewards on two specific games. It should be noted that this is an improvement over mere imitation; the policy can be better than those that were demonstrated by a human expert. Although RL and IRL in particular are in relative infancy within electron microscopy, they offer hope of automating tricky, multiple-optimum tasks that can be difficult to specify directly.

**IV.2.3. Active learning**

The discussion above would be incomplete without explicitly mentioning the possibility for the AE algorithm to query additional sources of information during the experiment, whether a theoretical model, published data, or even the human operator. As an example of the former, a ML method can be augmented by real-time modeling, including models based on the prior measurements (e.g., in model-based reinforcement learning), and models assisted by theory. In this case, the modeling that normally follows the experimental measurements becomes a part of the AE. This is a unique advantage of the AE, since human-based modeling is often associated with extraordinary large latencies. However, in this case, the key issue becomes the extent to which theory can reproduce the observed experimental situation. In other words, while experiment yields a certain representation of physical reality, so does a theoretical model. However, modeling results are also associated with certain representational biases and establishing matching between model predictions and experimental observables is a problem of its own (see Section VIII.4).

As a relevant parallel, we bring up autonomous driving. Here, the experimental effort is equivalent to the operation of a car based on immediate local information and the state of the road; yet the general route or presence of (known) alternative roads are not necessarily apparent to the driver. Comparatively, theory (in this case a map) provides the overall topology and connectivity of the travel space; yet theoretical modeling alone does not provide sufficiently exact information to stay on the road or to avoid other vehicles and obstacles. Hence, the co-navigation approach in this case maintains the car on the road while updating the map (e.g. discovering new roads and positioning known ones precisely) during driving.

Similarly, ML can use other sources of information, e.g., search for specific materials parameters, data mining, etc. An example of this process can be the search for a solid solution that has a particular dielectric value. In this case, the experiments provide information on the precise compositions of a phase diagram and the theory evaluates those for properties to feedback to the AE to ultimately find the best materials. Another example can be the design of the best single atom catalyst to drive an oxidation-dehydrogenation reaction, where theory would couple to AE which identifies the atom type and position in/on a support and theory would compute, say, the local density of states, which in turn might allow prediction of reactivity towards a particular system targeted for the reaction. In this case, the AE becomes closely connected to the issue of automated expert systems, etc., and is described in a number of other sources. This can be connected back to Bayesian methods presented earlier. In particular, flexible Bayesian methods are useful for indicating *disagreement* in the posterior distribution, which can be settled through external information like human labels.[98]

## IV. Microscope tuning

Operation of modern microscopes includes probe/imaging condition optimization as an inherent part of the analysis workflow. Unlike experiment planning that requires specific knowledge of material history and the general physical context in which an experiment is being performed, microscope tuning can often be performed without specific sample knowledge, using certain indicators of the probe behavior and calibration samples. This makes it an ideal target for machine learning based optimization.

### IV.1. Tip calibration/resolution optimization in SPM

Controlling the SPM/STM to achieve a probe state for high-resolution imaging is a time-consuming, very repetitive process that is well suited for reinforcement learning (RL)-based automation. The goal is to use RL for automated learning of complex relationships that are difficult to define except through trial-and-error. The main idea is rather straightforward: one uses a pre-trained deep neural network to categorize the probe state as "good" or "bad" and if the probe is categorized as "bad", the RL is used to learn a sequence of conditioning actions for obtaining a "good" probe. The examples of possible actions are scanning at high/low voltages and tip pulses of various length and amplitude. Recently, a proof-of-concept RL-based STM tip conditioning for a model system of MgPc molecules on Ag(100) was demonstrated.[99] In this case, the RL approach was based on a double $Q$-learning algorithm with an $\varepsilon$-greedy exploration strategy and it outperformed a random selection of actions (the more or less common way a STM operator approaches the tip conditioning problem) by ~28%.

In addition to RL-based strategies, one could utilize the GP-BO for tuning imaging parameters. A simple example would be utilization of the GP-BO for enabling the automated search of the optimal imaging conditions in the STM imaging parameter space (setpoint tunneling current, sample bias, etc.). In this case the specific tasks will include the choice of an optimal resolution function (defined as fast Fourier transform peak magnitude, boxcar averaged contrast variation on the atomic scale, or their combination) and risk functions defining the boundaries of the space with "safe" imaging. Here, as discussed above, the key part is the choice of the GP kernel and acquisition functions that severely affects the rate and efficiency at which the parameter space can be explored. Finally, one may substitute a "black-box" GP approach with a physics-based probabilistic model to account for interactions between the tip and surface and the contribution of different tip orbitals to the registered tunneling current, which may significantly alter the actual local density of states.

**IV.2. STEM tuning**

The resolution of an optical instrument is fundamentally limited by diffraction, which depends on the imaging aperture size and the wavelength of radiation used. The resolution limit in electron microscopy used to be incredibly important, not just because of a general desire for better images, but because the resolution limit tended to be just slightly worse than needed to reliably resolve single atoms and common atomic spacings other than in a few special cases[100, 101]. Because the wavelength of high-energy electrons is very small (a few picometers), the resolution limit is largely due how large an aperture angle can be used. Since the invention of the (S)TEM the aperture size has been constrained by the aberrations (imperfections) of the electron lenses. Thus, one of the most scientifically important developments in (S)TEM was the successful employment of aberration correctors about two decades ago[102-104]. However, although the general concept of how to correct aberrations was understood since Scherzer's work in the first half of the 20$^{th}$ century[105, 106], the key roadblock in implementing these systems was their complexity. An aberration-corrector consists of multiple stages and many of those stages need an alignment precision that is greater than can be achieved mechanically[107, 108]. These constraints mean that they require lots of adjustable parameters, each of which needs to be set accurately and then remain stable over the experiment duration. In practice, the settings tend not to be very repeatable, because of effects like drift, hysteresis, and temperature changes, which makes tuning and focusing such an instrument an ongoing challenge.

Consequently, the presently successful commercial aberration-corrector designs have integrated computer control. This control involves an analysis step that obtains an image, or sequence of images, which is used to diagnose the current status[102, 109, 110]. This information is then used to adjust the settings of the instrument. Clearly this process includes prior knowledge and implicit models, such as how the images relate to the aberrations, and how the settings should be altered to improve matters. In addition, there is usually some degree of flexibility to account for local variability. Each machine can either be manufactured differently or operated in new modes. For example, most microscope operators will be familiar with operations like setting pivot points, purifying a control, or calibrating the magnification.

Of particular relevance to the present discussion, there is some structural similarity between a simple neural network and the tuning software. Fundamentally the control scheme consists of a hierarchy of layers. The control process needs to translate the measured state to the desired change for a number of power supplies connected to the microscope. For a specific example, consider the case where the diagnostic software might indicate some undesired aberration. The first question is what to adjust – should it be fixed directly, or perhaps does it indicate some other misalignment? Once that issue is dealt with, both the magnitude and direction of the measured aberration then need to be translated into which controls to adjust. The controls could typically be a combination of magnetic poles in the appropriate planes, to match both the strength, symmetry, and direction, and then perhaps some combination of nearby alignment coils to keep the beam on track. Other controls might be even more complex, such as switching to a different imaging mode, then fixing a cascading multitude of different alignments.

Taking a further step back, there are underlying models included as prior knowledge: this process relies on the usual series of (S)TEM imaging approximations; the sample meeting certain conditions; and some modeling of the instrument to determine which controls to adjust, which was probably done during its design process[108].

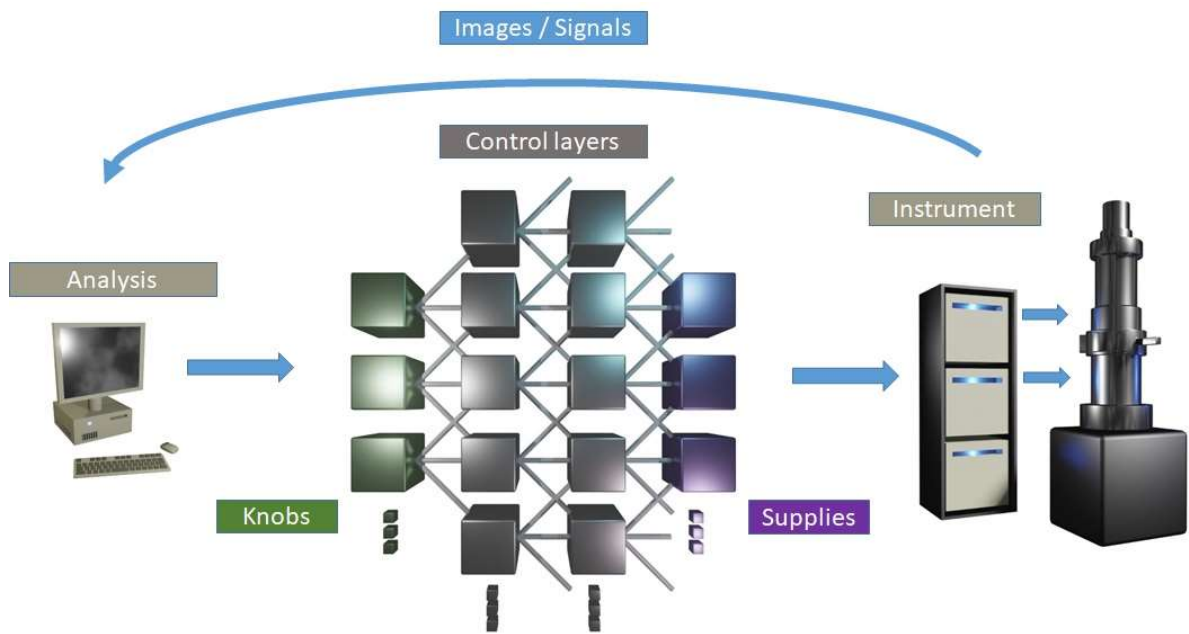

**Figure** (placeholder). Simplified cartoon schematic of the control system in a complex (S)TEM instrument with an aberration corrector. Images from the instrument are used to determine the current state, which is then used to decide which controls to adjust, which in turn affect the state of the instrument.

As a final note in this section, it is important to understand that most of the descriptions used here are already a deliberate over-simplification / model. In STEM we typically approximate the aberrations as a function over angle, however there are a lot of other factors that we neglect, such as energy (chromatic aberrations) and off-axial terms, or limited coherence (source size).

There are also unexpected factors like instrument malfunctions or mechanical issues. Recognizing when the automated procedure does not have enough information to proceed can also be important.

### IV.3. Image condition optimization in STEM

Thus far, we have divided the parameters to get a useful image in STEM into several broad categories. We have briefly mentioned the question of determining a material-of-interest, outlined setting some of the microscope parameters, and we now consider a third class of variables that concerns the condition of the sample. A lot of variables can impact the results, such as material thickness, position and orientation. For *transmission* electron microscopy, the sample usually needs to be prepared as a suitably thin film to allow most of the electrons to pass through, meaning that a good sample preparation facility is an essential pre-requisite for most electron microscopy.

The first step of many imaging sessions is usually a lower resolution survey. A prepared sample is typically on a 3mm diameter grid, but usually only a small area fraction is of interest. For materials/physics there are commonly two main classes of sample: either large crystals with some particular orientation; or small randomly oriented particles. In the former, the goal will be to locate an area of interest and tilt the sample onto a particular axis. For small particles, the goal might be to find some specific particles. (There are several exceptions to this statement, for example relatively large biological objects, or polymers, or essentially amorphous materials.) This survey process can be slow: for technical reasons, the high-resolution and low-resolution survey images might need microscope mode switches; and because of the fundamental difficulty of locating a very small object on a relatively large area. An obvious potential application of AE in this case is automatically finding the thing of interest.

Another useful example is sample tilt. In many cases accurately aligning the correct crystallographic axis with the microscope viewing direction is important for obtaining an interpretable image. There is a clear opportunity to automate this step, however, exactly how the angular information (such as a diffraction pattern) relates to a particular structure is a similar problem to imaging in real space. One relevant development in this area was the K-Space Navigator[111], which was developed because a new stage design made manually adjusting sample tilt more complex. The pertinent aspect of that system is that it implicitly contains crystallographic knowledge, and provides other assistance to the operator, such as automatically tilting to known directions. Such developments highlight an important first step in the implementation of more complete AE systems.

## V. Adjustable spectroscopy

The unique aspect of electron and scanning probe microscopy techniques is the availability of a broad range of spectroscopic modes. In these, a local signal is detected in the certain parameter space, providing information on local functionalities. The paradigmatic examples of spectroscopic imaging modes are electron energy loss spectroscopy (EELS) in STEM, current imaging tunneling spectroscopy (CITS) in STM,[25] and force-distance curve measurements in atomic force microscopy. Particularly broad range of spectroscopic methods have been developed in the context of Piezoresponse Force Microscopy[11, 112, 113] and current imaging, in which signals can be measured as a function of modulation bias, static bias, and time, giving rise to complex multidimensional data sets.[114-118] Until now, the spectroscopic measurements were performed in two preponderant regimes, namely point spectroscopy, in which the measurement is performed at a specific operator-selected location, and a grid mode, where the spectroscopic data is acquired over a predefined spatial grid. Parenthetically, the early introduction of machine learning methods in microscopy,[119, 120] starting from the linear unmixing methods such as Principal Component Analysis (PCA) and Non-negative Matrix factorization (NMF),[121-123] neural networks,[124] pansharpening,[71] direct datamining of structure-property relationships,[125, 126] etc. were driven by the necessity to analyze these data sets.

In many cases, the spectroscopic measurements involve sequential scanning of the parameter space and are associated with much longer signal acquisition time direct imaging. Correspondingly, hyperspectral measurements often require a balance between the number of spatial locations and sampling of the parameter space to enable reasonable data acquisition times. Hence, of interest is the development of AE methods for spectroscopic imaging. However, this requires both engineering controls that allow dynamic optimization of the spectroscopy waveforms and low latencies for ML algorithms.

### V.1. Image space

The most obvious strategy for the optimization of spectroscopic imaging is the collection of the data on non-gridded spatial locations. The identification of the regions of interest can be based on prior experiments over the same region, e.g., structural and functional images. For example, the specific domain walls in the PFM image or regions with specific tilt or curvature can be selected for spectroscopic measurements.

Alternatively, the experiment can be done similar to classical GP methods. Here, the spectroscopic measurements in several locations are used to get an initial set of responses, and GP can be used to identify subsequent locations. The key question that emerges in this case is the construction of the acquisition function. This can be based on the specific *a priori* defined parameter, e.g., area under the hysteresis loop in piezoresponse microscopy. Alternatively, it can be defined using the pre-trained transform, e.g., principle component analysis (PCA) or non-negative matrix factorization (NMF) component with known transform and DCNN or a (V)AE network. In both cases, the prior knowledge of the human in selecting the feature of interest, either via feature engineering or through a pre-trained network, forms the inductive bias imposed on the AE.

## V.2. Parameter space optimization

The second opportunity for spectroscopic measurements is associated with changing the measurement protocol itself. Here, the key factor is the physics of the measurement process, i.e., which parameters can be sampled in parallel (and if there are control options for assigning the detection bins) and which are sampled sequentially. Notably, major developments in microscopy are often associated with the transition from the sequential to parallel detection, with examples as diversified as Fourier Transform InfraRed spectroscopy (FTIR) in optical imaging and Band Excitation[127, 128] in SPM. Finally, the time domain can be sampled only within the duration of the experiment. Below, we discuss some of the opportunities for custom spectroscopy modes in SPM and STEM. In principle the spectroscopic measurement modality (e.g., detection frequencies) can also be changed from point to point based on the imaging data, although this has not been extensively explored.

### V.2.a. Spectroscopy in SPM

Scanning probe microscopy techniques have given rise to a broad spectrum of force, voltage, and time dependent spectroscopic measurements. These developments have particularly advanced in the field of Piezoresponse Force Microscopy that allows for 3-, 4, and 5D spectroscopies. In PFM, the basic signal is quantitative and independent of the tip-surface contact area,[129-132] allowing for direct interpretation of the spectroscopic data.[133] Similarly, a broad range of spectroscopies have emerged in the context of current voltage measurements, including 3D IV measurements in AFM and CITS in STM, and 4D first-order reversal curve imaging in conductive AFM.[134-136] Despite this broad range, in virtually all cases the spectroscopic measurements have been performed with a predefined probing waveform.

However, even a brief assessment of existing modalities for spectroscopic imaging in SPM suggests a tremendous range of opportunities for automatic and reconfigurable spectroscopies. Currently, many commercial and home-built systems allow for set-point detection for force and current measurements, in which the linear parameter ramp is stopped once the critical value of detected signal is achieved. This approach is typically used to obviate probe or sample damage during measurements. Similarly, it is straightforward to implement a measurement time threshold dependent on the changes in the detected signal.

Broad opportunities exist in the context of multidimensional current- and voltage spectroscopies in the context of electromechanical (Piezoresponse Force Microscopy, Electrochemical Strain Microscopy) and conductive measurements. In this case, the phenomena of interest are distributed non-uniformly in the voltage space, with domain nucleation corresponding to well-defined nucleation biases. Correspondingly, uniform sampling of the multidimensional parameter spaces as performed in variants of PFM spectroscopies is deeply sub-optimal, and development of algorithms that effectively sample the regions of interest is a high priority. It should be noted that efficiency and veracity of these algorithms can be greatly enhanced once the physics-based model is available, e.g., the Preisach model for polarization switching or equivalent circuit model for IV measurements.[137]

**V.2.b Spectroscopic imaging in STEM**

In (S)TEM the electrons that are used to form the image also interact with the sample in various ways. In particular, as they pass through the sample, these high-energy electrons transfer energy to the sample[138, 139]. This electron energy loss signal (EELS) can be quantified in two main ways: By using a spectrometer to measure the relative kinetic energy of the fast electron; or as the electrons inside the sample relax back to their previous state they release the excess energy as photons, which can be detected with cathodoluminescence or X-ray systems.

One fundamental problem with detecting these signals is that the amount of energy lost is usually relatively small and that the loss-probability decreases sharply with the amount of energy transferred. For example, for new monochromated STEMs we might want to find ~meV losses on a ~100keV beam; and the signal of interest is frequently only a small fraction of the total number of electrons[140-143]. To address these problems, an obvious application for AI/ML is denoising – both boosting the interesting signal relative to the noise and removing acquisition artifacts. One common way to address this issue is based on surrounding spectra, prior knowledge, or theory, to establish a model of how we would expect this spectrum to behave. However, some caution is needed. Imagine the case of a single dopant signature in a large crystal – the question might be to decide if an unexpected signal is statistically significant. This process is already commonly performed, although usually only in post-acquisition analysis. On the fly decision making in AE opens up new possibilities: Can we go back and repeat an unusual measurement? Similarly, if we recorded a few nearby spectra or used different conditions, could that boost our confidence in the result? Finally, if the data is very sparse, can we efficiently compress it at the data generation point?

One area of current interest is using the EELS signal to measure magnetic properties by comparing subtle changes in the shape of the signal[144-146]. The biggest problem is that this signal tends to have a really small signal difference between two already small signals. Clear applications of AE would include noise reduction or concentrating on an interesting signal. A related method was demonstrated by Idrobo et al.[147], based on the theoretical development of Rusz[148], where the symmetry of the electron probe was used to amplify the magnetic component of the EELS signal. Essentially this experimental result consisted of tuning the acquisition conditions aberrations to match the symmetry of the crystal, which is perhaps the sort of route that AI/ML might be better suited to follow than a human operator.

Another complication, highlighted by the inclusion of the EELS signal, is that the data is inherently multidimensional. Adding an energy-loss dimension for every position in a two-dimensional (2D) image results in a 3D spectrum image. Speaking generally, the energy of a particle depends on its wave-vector, meaning that in many cases the directional dependence of the energy loss function will also be interesting. In other words, we frequently might want to record the energy-loss as a function of scattering angle. Adding in the two scattering directions available in the (S)TEM geometry leads to a 5D angle-resolved EELS dataset. Given that even todays most advanced detectors are typically only 2D means that there is no obvious method to record all of these dimensions simultaneously. Similar to how a STEM probe is scanned over position to record a 2D image, we might want to scan over some of these other dimensions. However, scanning over multiple dimensions will be time-consuming, meaning that the experiment might effectively be impossible due to drift or dose-dependent damage. In this case, an AI/ML based method capable of deciding what part of this high-dimensional dataset to acquire could be a vital component.

Another simple aspect of AE might be to just automate some of the tedious repetitive nature of many experiments. Consider the general case where the goal is to get multiple scans over the same particle or area of interest, while varying acquisition conditions. A good example of this process, for which some ML has already been partly demonstrated, is tomography. In this technique, multiple images are obtained at different tilts, thereby allowing reconstruction of a 3D model. One of the usual practical problems is that when the tilt is changed, the area of interest moves. Finding and refocusing on this area can be difficult. There is potential application for simple ML in establishing a predicted translation for a given tilt, which can then be combined with a search routine to automatically locate and focus on the selected object. Another common task in which ML has only been partially exploited is how to infer 3D shapes from a series of projections. The same goal might be achieved for a through-focus depth series, changing focus to optically section through a sample. Or by some combination depth + focal series, especially in the dose-limited case where every image has to be used, whether it is obtained at the desired conditions or not.

In passing we mention the so-called 4D STEM or ptychography, where a 2D electron scattering pattern is recorded for every position in a 2D image scan. While this dataset is obviously far larger and can be complex to interpret, it possesses many potential advantages. For example, as demonstrated by Rodenburg and Nellist[149-152] that the usual image resolution limits do not apply in the same way; and that it is extremely sensitive to the local field distributions[153, 154]. These datasets are usually highly redundant and currently reconstructions are obtained through time-consuming iterative methods, meaning they hold vast promise for the application of AI/ML in future.

## VI. Beyond rectangular scanning

Finally, the discussion of the automated experiment will be incomplete without considering the potential and opportunities offered by non-rectangular scans. Currently, both the SPM and STEM fields almost invariably rely on classical rectangular raster scans in which the beam or the probe rapidly traverses the surface in the fast scan direction, and slowly shifts in the perpendicular direction forming the slow scan direction. This scanning mode offers both the advantage of easy implementation and yields data in the form of 2D maps that can be readily interpreted by a human eye. However, it is also associated with significant and well-understood limitations. For example, in STEM the rectangular scanning is associated with distortions due to the fly-back effects, where the beam motion lags behind the rapid change of nominal position due to the capacitive/inductive effects in the control circuits. Similar problems emerge in the context of fast scanning in SPM due to the inertial effects of the scanner. In several cases, scanning following spiral or Lissajous curves has been proposed and implemented to alleviate these issues.

Similarly, beyond rectangular scanning becomes a key prerequisite for the AE aimed at structural discovery, minimizing the surface damage, or attempting controlled modification of the surface. The possible paradigms for AE in this case are summarized in Figure 7. Ultimately, we can envision the freeform scanning approaches, where the direction and velocity of the probe motion are updated based on previously detected information. However, given the latencies of SPM and STEM imaging, this will necessitate development of specialized light algorithms and edge computations, as discussed below. On a shorter time frame, adapting the parameters of the predefined waveform, e.g., pitch of the spiral or line density in rectangular sub scans offers a more practical alternative.

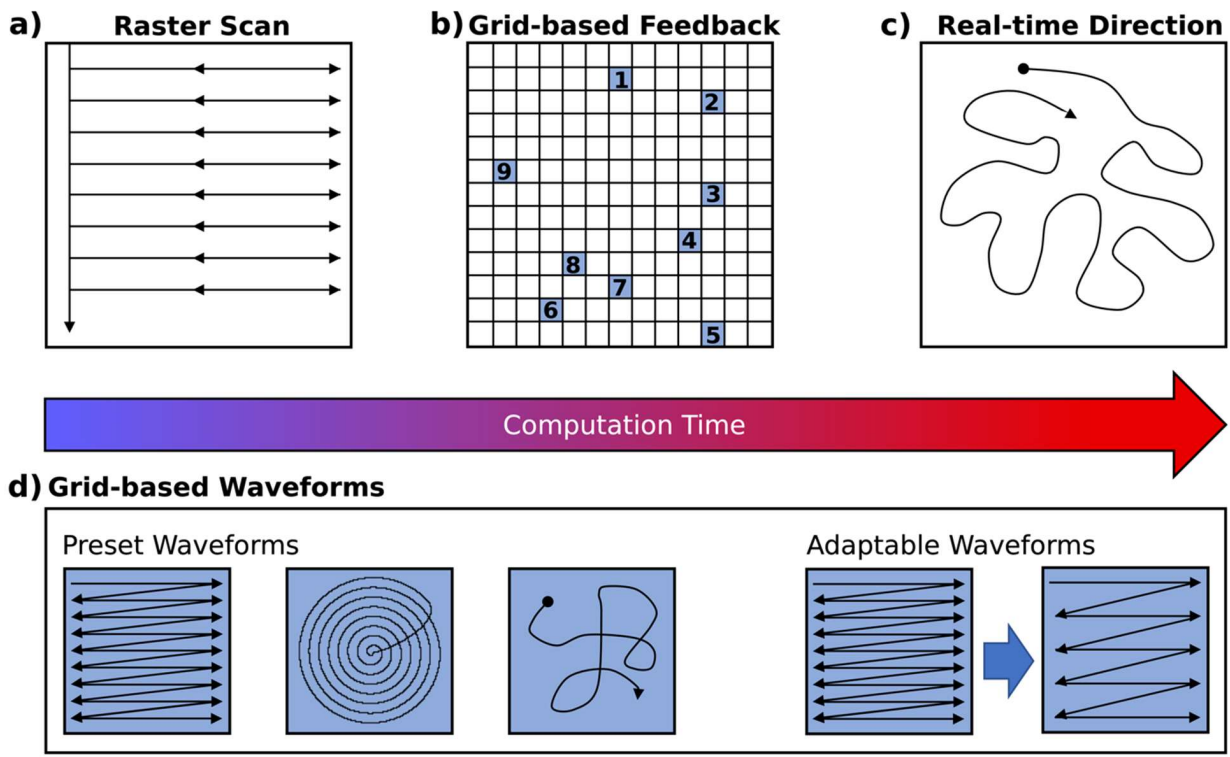

**Figure 7.** Possible scanning paradigms in SPM and STEM. Shown are (a) classical scan, (b) sub-scan-based images for structural discovery, based either on preset or adjustable scan-forms shown in (d), or (c) freeform scanning. The selection of image regions in (b) or scanning direction and velocity in (c) are guided by a suitable ML algorithm.

The adoption of non-rectangular scanning will further necessitate two specific developments. One is the reconstruction of the classical 2D images from the non-uniformly sampled data. Currently, the approaches for such reconstruction include GP [155] and Compressed Sensing,[156-159] and many other algorithms can be envisioned. The second practical consideration includes drift correction. While many commercial systems are optimized to compensate for the scan non-linearities and minimize drift, the adoption of non-rectangular scan paths will necessitate either dynamic compensation of drifts or higher stability of measurement systems, or incorporation of drift compensation as a part of the workflow.

**VII. From AE to fabrication/nonstationary of the sample**

Finally, as a special (but potentially the most important) case of automated experimentation is the AE where a material evolves under the action of external stimuli or is modified under the action of the scanning probe or electron beam. In the first case, the task of the AE is to acquire most information when materials structure changes. In the second, we can envision applications where the aim of an automated experiment is to learn the cause and effect relationships between the local action and material response, and harness these towards the creation of specific structures.

**VII.1. Dynamic phenomena induced by external stimuli**

Direct observation of dynamic phenomena in scanning probe and electron microscopy is broadly considered as a pathway to explore the physics and chemistry of the relevant systems. Multiple examples include liquid-cell electrochemical studies in STEM and SPM, piezoresponse force microscopy of polarization switching in ferroelectric films and capacitors, temperature induced phase transitions, and catalytic processes. These transformations can be induced by global thermal or chemical stimuli, fields applied via microelectrodes, external probes, or by the imaging probe. For reversible processes possessing local return point memory (meaning that each region of parameter space corresponds to a well-defined microstructure independent of the path), the AE can be performed in the generalized parameter space including both spatial and control variables, with the appropriate pathfinder function that balances the latencies of spatial and control parameter exploration. For irreversible processes induced by global stimuli, the AE paradigms are limited and perhaps only the rate of the control variable change can be varied dependent on the transformations. For example, variable temperature experiments can proceed with a rapid heating rate when the materials structure does not evolve and then slow down at the phase transition temperatures.

Finally, we believe the most important application case for AE in SPM and STEM is the case when the probe locally introduces an irreversible, or partially reversible, changes in a material. Multiple examples of such behavior include local wear and nano-oxidation in SPM experiments and beam damage in electron microscopy. While generally considered to be a problem for imaging, these phenomena offer a clear opportunity for material modification on the nanometer and even atomic level. Harnessing such opportunities in turn necessitates understanding the cause and effect relationships between local action (for example, tip bias in SPM, or dwell time, dose, or electron energy in STEM), and resultant material changes.

A trivial example (from a ML perspective) is offered by the cases where the cause and effect relationship is deterministic and can be discovered experimentally, providing the full information for experiment planning. For example, in tip-induced nano-oxidation of metals and semiconductors[9, 160-163] or electrochemical writing in lanthanum aluminate-strontium titanate interfaces,[164, 165] once the critical tip bias is established, complex nanostructures can be created. Similarly, an electron beam can be used to crystallize or amorphize semiconductor materials[17, 166-170] or induce polarization switching in ferroelectrics.[18, 19] Once the relationship between electron dose and the resultant structures is established, a controlled path scanning can be used to create structures of predefined spatial complexity.[20] A recent overview of such an effort is given in Ref. [31]. Note that this approach is then similar to the mainstay lithographic methods in that the known cause and effect relationships are used to define the process (and finding this relationship forms a significant part of the semiconductor process development). Note that the approach can extend to the atomic level, as exploited by the STM based atomic fabrication that is now used as the technological basis for qubit formation, and recent demonstration of controllable motion of Si atoms in graphene.[22-24]

Comparatively, the electron beam is known to induce a broad spectrum of transformations including vacancy formation, diffusion, ordering, phase transformations, formation of extended defects, etc. Generally, these processes are sensitively dependent on the beam energy, potential presence of the contaminants on the surface or vacuum system and can be highly materials specific. Recently it has been shown that beam can introduce the broad range of single-atom changes, including vacancy formation, single atom dynamics, and formation of defects. The most important aspect of these processes is that they are often stochastic. Correspondingly, implementation of ML strategies for harnessing these mechanisms for atom by atom fabrication presents a clear opportunity.

**VIII. What is needed for AE**

The discussion above illustrates potential considerations relevant to the design and implementation of AE workflows in (S)TEM and SPM, specifically addressing the issue of prior knowledge and inductive biases in the ML element. However, the practical implementation of the AE workflows also requires a solution of several engineering problems.

**VIII.1. Engineering controls**

The first and most significant bottleneck in the implementation of the AE is the instrumental controls. Most of the commercial tools to date had predominantly closed control systems, with some functionality only available only to human operator. This is partially due to risk minimization (it is undesirable for a software bug to cause physical damage, or to create a potential safety hazard), and often stem from specific monetization schemes where the instrument service and low-level modifications are deferred to company representatives. A relatively small, but growing, number of SPM and STEM manufacturers now support scripting or allow more low-level controls of the tools.

That said, the instrument level controls represent the first and most significant bottleneck in AE, defining the space of possible in experiment design. Hopefully the future will see the emergence of multi-tiered control schemes at different risk levels, and provide instrumental simulators, and tools that allow assessment of risks prior to experiments. Recent advances in fields such as Arduino or Raspberry Pi based systems, or internet of thing (IoT) electronics, suggests that the development of such infrastructure is entirely possible.

**VIII.2. Edge computing**

The second key aspect of the AE in microscopy is the need for edge computing, i.e., deployment of the ML algorithms close to the point of data generation. Classically, ML methods are implemented either on local resources such as computers and workstations with GPUs, local clusters, or cloud services belonging to the organization or provided by Google, Amazon, Microsoft, or other commercial vendors. For the vast majority of applications (except for financial markets or autonomous driving), the short-term latencies of ML methods are not particularly relevant and the time for training and prediction can be readily accommodated. However the situation is totally different for the AE in microscopy. Here, the relevant time scales are set by the hardware limitations. For scanning probe microscopy, the typical frame time is ~10 min, corresponding to 1 scan line/second. For (S)TEM, 10 ms – 100 s per image acquisition rates are all common, depending on the conditions and desired signal to noise ratio. Finally, in both SPM and STEM there is a considerable interest towards fast measurements, corresponding to video-rate imaging for SPM, and with examples such as few ms frame rate in TEM. For spectroscopic measurements both in SPM and STEM, several minute time scales are common. Correspondingly, implementation of the AE necessitates operation of ML algorithms at these time scales.

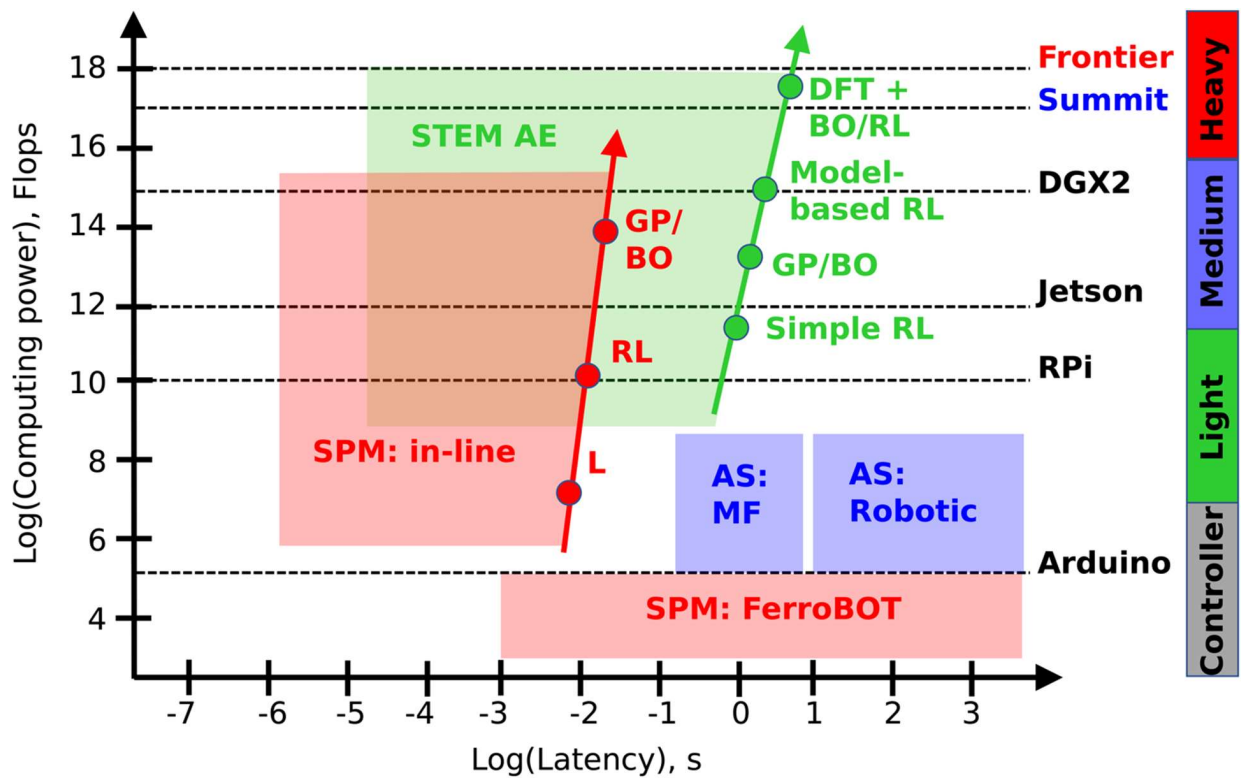

**Figure 8.** Automated experiment in (scanning) transmission electron microscopy holds the promise of quantitative and high-veracity imaging, computer-driven rapid characterization workflows, and electron beam based atomic assembly. Realization of these opportunities

necessitates implementation of machine learning algorithms that can guide experiments at the intrinsic latencies of image acquisition. This figure semi-quantitatively illustrates the estimates for the latencies and required computational capacity for the implementation of the various ML algorithms including GP/BO and model-based and model-free reinforcement learning, and the computational capacity of edge devices available currently. Here, Frontier and Summit refer to the capabilities of ORNL supercomputers.

A roadmap for the latencies and required computer power for different ML algorithms is shown in Figure 8. We note that simple line-by-line feedback operation as incorporated in the FerroBOT[94] can in principle be realized on very light edge computing devices, even such as Arduino. The Raspberry Pi and Jetson series from NVIDIA allow for deployment of the pre-trained deep learning models with the required latencies. Finally, the implementation of the GP/BO algorithms (without optimization) typically requires higher-level computational capabilities.

Two considerations should be mentioned going forward. First, broad propagation of the AE workflows will lead to the rapid progress in light ML algorithms that can run on limited hardware. These will likely be based on similar development for devices such as cell phones, wearable electronics, and various IoT sensors. Secondly, an important consideration for some imaging modes (e.g., 4D STEM and varieties of X-Ray imaging and tomography) will be the transfer of potentially enormous volumes of data.

**VIII.3. Simulators and emulators**

One critical element of broad incorporation of AE workflows in (S)TEM and SPM community will be the development of more advanced instrument simulators. Currently, typical SPM tools are priced in the 100-300 k$ range, with the prices of ultrahigh vacuum machines being in the 1 – 2 M range and STEMs in the 2-5M$ range. These machines often operate as a part of large user facilities, with relatively small-time windows available for maintenance and upgrades. Combined with the risks of trying something new (costs and time for repair), this makes direct implementation of AE workflows on the operational instrument impractical (with possible exception of ambient SPMs). Correspondingly, a crucial part of the development of the AE workflows is the broad dissemination of the microscope simulators, ranging from the simple black box devices that can stream stored data at the realistic imaging speeds to simulators that contain a model of the materials properties and an imaging simulator, and thus can simulate the effects of the microscope controls on images. Broad dissemination of these will significantly accelerate the AE progress by introducing a broad range of scientists to microscopy. Here, additional acceleration of the progress can be achieved via equivalents of Kaggle competitions, etc.

**VIII.4. Theory-experiment matching on the fly**

As mentioned above, matching theory with experiment is not always simple due to intrinsic assumptions in theoretical models (structure, defects, approximations in force-fields or wavefunction, lack of using a realistic environment such as a solvent, time and length scale mismatches, etc.) as well as difficulties in precision determination of relevant comparators in experiments.[171] Theory will always give an answer, but without validation, it can be over idealized and potentially misleading. The key to enable true predictive power is feedback with experiments to optimize the overall process such that meaningful results are obtained. For example, for solid crystalline materials, experimental structures are precise and even provide accurate information down to the single atom level as well as their time evolution. This information which can be obtained via scattering, spectroscopy and microscopy provides precise coordinates and atomic compositions for a theoretical model which would match the experiment at the given time of the experimental measurement. However, one must establish a workflow for the data and imaging analysis to provide the relevant atomic configurations, as well as response behaviors of materials for direct input into first principles simulations, and to enable subsequent refinement of theoretical parameters via iterative feedback. This information can then allow a quantum-based calculation such as density function theory to access properties (electronic, mechanical, etc.) as well as to evolve the dynamics of the system.[43] We note that true materials are far more complicated than the simple structures often studied in small periodic unit cells by electronic structure calculations. The vast length and time scales, as well as finite temperatures, make even density functional theory (DFT) calculations for investigating materials under device-relevant conditions prohibitively expensive. Grain boundaries, extended defects and complex heterostructures further complicate this issue. Evaluating defects or diffusion of atoms that can be compared to experiment via time resolved experimental measurement[172] but that also faces strict concern due to the time scale of resolved measurements which is often much longer than that which can be computed via theory. In this regard ML surrogates can help bridge that gap, again being calibrated by high-level theory and experiments to ensure accuracies.[173-176] This also can help make on-the-fly matching a possibility as compared to traditional theoretical approaches which would require off-line calculations due to the computational complexities.

Finally, we note that microscopy tools can be used to arrange atoms in desired configurations, thus effectively controlling matter via current- and force-based scanning probes and electron beams.[20, 24, 177] This opens a pathway to explore non-equilibrium and high-energy states generally unavailable in the macroscopic form but often responsible for materials functionalities. Here the feedback with theory will be important to guide the experiments to the correct locations and perturbations to enable precise sculpting of a material to achieve a desired structure or property. In particular, at the highly focused and localized energies typical under a STEM, atoms can be excited to electronic states beyond the ground state. This means one needs to consider coupled nuclear-electronic dynamics in order to fully evaluate what potential energy landscape an atom experiences in the experiment.[178] Such calculations are difficult but now possible and have proven to be matched well to experiment. Thus, building a ML surrogate becomes possible and should help enable near real-time feedback with experiment. This can therefore complete a full loop in the materials discovery and design cycle, from exquisite observation to theory-based prediction, and finally to experimental control of materials.

In closing, we note that one of the more enabling aspects of theory–experiment matching is to potentially improve a theoretical model based on experimental observations. For example, on the qualitative imaging provides direct observation of atomic configurations that gives more information on local structures, defects, interfaces and so on. On a semi-quantitative level, the numerical values of observed atomic spacing's can indicate the incompleteness of a model, e.g., the presence of (invisible) light atoms or vacancies.

## IX. Summary

Automated and autonomous experimentation (AE) microscopy is attracting a progressively larger share of attention from the scientific community. Here we argue that that fully autonomous workflows lie in the domain of the distant future, if at all possible, since microscopy experiments are well defined only in the context of prior physical knowledge and its usefulness is measured in the context of the additional information it discovers. However, the automation of experiments in microscopy is possible once the physical priors and inductive biases available to the AI/ML system prior to the experiment are well understood and the information transfer between the AI/ML control and the outside knowledge pools are defined. This analysis allows us to set clear expectations for AE in microscopy.

Practically, we believe that broad incorporation of AE in microscopy will proceed via automatization of the specific bottlenecks in the microscopy workflows, including probe conditioning, guided exploration of large images or optimized spectroscopy measurements. The initial role of AI/ML will be to automate the time-intensive and repetitive operations, as well as to provide suitable human level inputs that allow human control of much faster imaging process at intrinsic microscope latencies.

We further note that currently the biggest bottleneck towards broad implementation of the AE in STEM and SPM are the engineering controls of the instrumental platforms that rarely allow the external controls of scanning, spectroscopy, and data acquisition. However, recent emergence of environments such as Nion Swift, JEOL PyJem, in STEM suggest that the situation is starting to rapidly change. In SPM, customizable controllers such as Nanonis and Asylum Research have long been available. Notably, even relatively low-cost edge computing platforms such as Arduino, Raspberry Pi, and NVIDIA Jetsons can allow sufficient latencies and computational capabilities to enable AE once the engineering connections are established and the relevant algorithms are available. However, we argue that existing ML tools may be insufficient to fully realize the AE potential, since the experiments are always associated with the uncertainties and observational biases that can lead to out of distribution data, etc. From this perspective, quantification (defined as transition from microscope observables to materials specific descriptors) becomes paramount. Similarly, another key is the incorporation of the physical models that provide regularized priors in the Bayesian and deep learning methods. Finally, many ML algorithms, especially the model-based ones, can be both slow and computationally intensive compared to typical data acquisition times. This in turn necessitates the rapid adoption of the edge computing capabilities as a way to provide the necessary computational power directly at the point of data generation.

Overall, we argue that role of the AE in microscopy is not the exclusion of human operators (as is the case for autonomous driving), but rather automation of routine operations such as microscope tuning, etc., prior to the experiment, and conversion of low latency decision making processes on the time scale of the image acquisition to the human-level high-order experiment planning. It is the second area that opens the most fertile ground for scientific exploration and ML developments. When adopted, ML/AI can dramatically alter the (S)TEM and SPM field; however, this process is likely to be highly nontrivial, be initiated by combined human-ML workflows, and will bring new challenges both from the microscope and ML/AI sides. At the same time, these methods will enable fundamentally new opportunities and paradigms for scientific discovery and nano- and atomic structure fabrication.

**Acknowledgements:** This effort (ML and STEM) is based upon work supported by the U.S. Department of Energy (DOE), Office of Science, Basic Energy Sciences (BES), Materials Sciences and Engineering Division (S.V.K., S.J., A.R.L.), by U.S. Department of Energy, Office of Science, Office of Basic Energy Sciences Data, Artificial Intelligence and Machine Learning at DOE Scientific User Facilities program under Award Number 34532 (A.G., B.G.S.) and was performed and partially supported (M.Z.) at the Oak Ridge National Laboratory's Center for Nanophase Materials Sciences (CNMS), a U.S. Department of Energy, Office of Science User Facility.